\documentclass[11pt,a4paper]{article}
\usepackage[hyperref]{emnlp2020}
\usepackage{times}
\usepackage{amsmath}
\usepackage{latexsym}

\usepackage{url}

\usepackage{graphicx}  
\usepackage{amsfonts}
\usepackage{CJK}
\usepackage{bm}
\usepackage{mathrsfs}
\usepackage{float}
\usepackage{xcolor,colortbl}

\usepackage{subfigure}
\usepackage{booktabs,multirow}
\usepackage{bigstrut,bigdelim}
\usepackage{paralist}
\usepackage{bbm}

\usepackage{helvet} 
\usepackage{courier} 
\usepackage{makecell}
\usepackage{nicefrac}       

\usepackage{microtype} 
\usepackage{epsfig}
\usepackage{diagbox}

\usepackage{framed}
\usepackage{empheq}
\usepackage{array}
\usepackage{enumitem}
\usepackage{mdwlist}
\usepackage{xspace,mfirstuc,tabulary}

\usepackage{tcolorbox}

\usepackage{algorithm}
\usepackage{placeins}

\usepackage{algpseudocode}

\newcommand{\thickhline}{\noalign{\hrule height 1pt}}

\usepackage{microtype}

\aclfinalcopy 
\def\aclpaperid{1006} 

\title{Knowledge-Grounded Dialogue Generation with \\ Pre-trained Language Models}

\author{
Xueliang Zhao$^{1,2}$, Wei Wu$^3$, Can Xu$^4$, Chongyang Tao$^4$, Dongyan Zhao$^{1,2}$, Rui Yan$^{1,2,5}$\thanks{Corresponding author: Rui Yan (ruiyan@pku.edu.cn).}~ \\
$^1$Wangxuan Institute of Computer Technology, Peking University, Beijing, China \\
$^2$Center for Data Science, AAIS, Peking University, Beijing, China \\
$^3$Meituan, Beijing, China \quad
$^4$Microsoft Corporation, Beijing, China\\
$^5$Beijing Academy of Artificial Intelligence (BAAI), Beijing, China \\
\texttt{\{xl.zhao,zhaody,ruiyan\}@pku.edu.cn} \\
\texttt{\{wuwei19850318,chongyangtao\}@gmail.com} \\
}

\date{}

\begin{document}
\maketitle
\begin{abstract}
We study knowledge-grounded dialogue generation with pre-trained language models. To leverage the redundant external knowledge under capacity constraint, we propose equipping response generation defined by a pre-trained language model with a knowledge selection module, and an unsupervised approach to jointly optimizing knowledge selection and response generation with unlabeled dialogues. Empirical results on two benchmarks indicate that our model can significantly outperform state-of-the-art methods in both automatic evaluation and human judgment.

\end{abstract}

\section{Introduction}

With advances in neural machine learning \cite{sutskever2014sequence,gehring2017convolutional,vaswani2017attention} and availability of the huge amount of human conversations on social media \cite{adiwardana2020towards}, building an open domain dialogue system with data-driven approaches has attracted increasing attention from the community of artificial intelligence and natural language processing. In this work, we are interested in generative approaches. Generative models for open domain dialogues are notorious for replying with generic and bland responses, resulting in meaningless and boring conversations \cite{li2015diversity}. Such deficiency is particularly severe when human participants attempt to dive into specific topics in conversation \cite{dinan2018wizard}.  As a result, there is still a big gap between conversation with existing systems and conversation with humans. 

\begin{table}[]
\footnotesize
\begin{tabular}{c|p{150pt}}
\hline
\multicolumn{2}{c}{Context} \\ \hline
\multicolumn{1}{c|}{A}        & I just discovered star trek and I really like watching star trek . \\
\multicolumn{1}{c|}{B}        & Gene Roddenberry created it based upon science fiction and it is American media. \\
 & ... \\
\multicolumn{1}{c|}{A}        & If I remember Captain Kirk was not the original captain . \\
\multicolumn{1}{c|}{B}        & The Star Trek Canon of the series an animated had 5 spin offs. \\
\multicolumn{1}{c|}{A}        & I watched a little of the next generation but could not get into it like i did with the original show . \\ \hline
\multicolumn{2}{c}{Response}  \\ \hline
\multicolumn{1}{c|}{Human}    & These adventures went on but were short lived and six feature films. \\ \hline
\multicolumn{1}{c|}{DialoGPT} &  I think it's worth it. \\ \hline
\end{tabular}
\caption{An example from the test set (Test Seen) of Wizard of Wikipedia \cite{dinan2018wizard} .}  
\label{tab:intro}
\end{table}

Very recently, there emerge two lines of research that seem promising to bridge the gap.  One is to apply large-scale pre-trained language models, such as GPT-2 \cite{radford2019language}, to the task of open domain dialogue generation. Prototypes such as DialoGPT \cite{zhang2019dialogpt} have exhibited compelling performance on generating responses that make sense under conversation contexts and at the same time carry specific content for keeping the conversation going.  While the giant language models can memorize enough patterns in language during pre-training, they only capture ``average’’ semantics of the data \cite{zhang2019dialogpt}. As a result, responses could still be bland or inappropriate when specific knowledge is required, as illustrated by the example in Table \ref{tab:intro}. The other line is to ground dialogue generation by extra knowledge such as unstructured documents \cite{zhao2020low}.  By the means, the documents (e.g., wiki articles)  serve as content sources, and make a dialogue system knowledgeable regarding to a variety of concepts in discussion. However, collecting enough dialogues that are naturally grounded on documents for model training is not trivial.  Although some benchmarks built upon crowd-sourcing have been released by recent papers \cite{zhou2018dataset,dinan2018wizard,gopalakrishnan2019topical},  the small training size makes the generation models generalize badly on unseen topics \cite{dinan2018wizard} and the cost of building such data also prevents from transferring the techniques proved on the benchmarks to new domains and new languages.

Encouraged by the results on pre-training for dialogue generation and knowledge-grounded dialogue generation, and motivated by the problems in both sides, we consider bringing the two together in this work. Specifically, we propose knowledge-grounded dialogue generation with pre-trained language models in order to endow a generative model with both rich knowledge and good generalization ability\footnote{In this paper, we assume that knowledge is retrieved from documents.}. The challenge is that pre-trained language models often set constraints on the maximum number of tokens they can handle (e.g., the maximum number for GPT-2 \cite{radford2019language} is $1024$), and thus hinders exploitation of the knowledge text which could be rather long and redundant (e.g., in Wizard of Wikipedia \cite{dinan2018wizard}, on average each conversation context is associated with $61.2$ sentences retrieved from wiki articles, and the average number of tokens in the extra knowledge is $1625.6$). Indeed, the conflict between model capacity and the ability required for processing long knowledge input represents an essential obstacle for applying pre-trained language models to knowledge-grounded dialogue generation, since on the one hand we always have to set up an upper bound to the capacity of pre-trained models in order to handle massive text corpus, and on the other hand we need to keep sufficient candidates with rich enough content in the procedure of response generation in order to guarantee the recall of relevant knowledge. 

To overcome the challenge, we consider equipping the pre-trained response generation model with a knowledge selection module whereby the redundant knowledge input is slimmed with relevant information (regarding to conversation contexts) kept to meet the capacity constraint. While some recent papers on knowledge-grounded dialogues have paid attention to the problem of knowledge selection \cite{lian2019learning,kim2020sequential,ren2019thinking}, the knowledge selection module is either deeply coupled with the specially configured models \cite{lian2019learning,ren2019thinking} and thus is incompatible with the pre-trained language models, or it is learned with human annotations \cite{dinan2018wizard,kim2018semantic} which are difficult to obtain in practice (e.g., the dataset in \cite{zhou2018dataset} does not contain annotations for knowledge selection). Therefore, we propose an unsupervised approach where learning of knowledge selection and fine-tuning of response generation are jointly conducted with unlabeled dialogues. Specifically, we build the knowledge selection module on the basis of BERT, and formalize knowledge selection as a sequence prediction process, by which the model can take advantage of the pre-training techniques and dynamically determine the relevant knowledge for a given context. The learning algorithm starts from training with pseudo ground-truth that is constructed by making full use of responses as an alternation of human annotations, and then alternatively updates the knowledge selection model and the response generation model through a reinforcement learning approach and a curriculum learning approach respectively. Thus, knowledge selection is further optimized with the feedback from response generation, and the knowledge used for fine-tuning the response generation model gradually moves from the pseudo ground-truth to the prediction of the knowledge selection module. 

We test the proposed method on two benchmarks of knowledge-grounded dialogue generation: Wizard of Wikipedia \cite{dinan2018wizard} and CMU Document Grounded Conversations \cite{zhou2018dataset}. Evaluation results indicate that our model can significantly outperform state-of-the-art methods as well as a few pre-trained models used in heuristic ways, and thus achieves new state-of-the-art on the benchmarks. Moreover, as a byproduct, the knowledge selection module also outperforms the state-of-the-art model in terms of accuracy of knowledge selection on Wizard of Wikipedia, implying that other models could also benefit from the component. 

Our contributions in this paper are three-fold: (1) proposal of a knowledge selection module for applying pre-trained language models to the task of knowledge-grounded dialogue generation; (2) proposal of an unsupervised approach in which learning of knowledge selection and fine-tuning of the pre-trained model are conducted in a joint manner; and (3) empirical verification of the effectiveness of the proposed method on benchmarks of knowledge-grounded dialogue generation.

\section{Related Work}
Early work on end-to-end open domain dialogue generation is inspired by the research of machine translation \citep{ritter2011data,shangL2015neural,vinyals2015neural}. Later, the vanilla encoder-decoder architecture is widely extended to improve diversity of responses \cite{li2015diversity,xing2017topic,zhao2017learning,tao2018get}; to model the structure of conversation contexts \cite{serban2016building,serban2017hierarchical,xing2017hierarchical,zhang2019recosa}; to control attributes of responses \cite{xu2019neural,zhou2017emotional,zhang2018learning,wang2018learning,see2019makes}; and to bias responses to some specific personas \cite{li2016persona,zhang2018personalizing}. 
Recently, grounding dialogue generation by extra knowledge is emerging as an important step towards human-like conversational AI, where the knowledge could be obtained from knowledge graphs \cite{zhou2018commonsense,moon2019opendialkg,tuan2019dykgchat}, retrieved from unstructured documents \cite{dinan2018wizard,lian2019learning,zhao2020low,kim2020sequential}, or extracted from visual background \cite{mostafazadeh2017image,shuster2018engaging,huber2018emotional}. In this work, we study document-grounded dialogue generation. Rather than learning from scratch like most existing work, we take advantage of the pre-trained language models and achieve new state-of-the-art on the benchmarks of the task.

 
 
Big, deep neural language models pre-trained on huge unlabeled text corpus have led to strong improvements on numerous natural language understanding and natural language generation benchmarks \cite{devlin2018bert,yang2019xlnet,liu2019roberta,radford2019language,song2019mass,dong2019unified,lewis2019bart}, and therefore are revolutionizing almost the full spectrum of NLP applications \cite{raffel2019exploring,sun2019utilizing,qiao2019understanding,zhang2019hibert,lample2019cross} and some 
interdisciplinary applications in NLP and computer vision \cite{lu2019vilbert,su2019vl,sun2019videobert}. In the context of dialogue generation, by fine-tuning GPT-2 \cite{radford2019language} in different sizes on social media data, recent work has \cite{zhang2019dialogpt,wolf2019transfertransfo} shown promising progress on conversation engagement and commonsense question-answering. In this work, we further explore the application of pre-training to the task of open domain dialogue generation by equipping the pre-trained language models with external knowledge. Different from a very recent paper on pre-training for low-resource knowledge-grounded dialogue generation \cite{zhao2020low}, the work presents an in-depth investigation on how to release the power of the existing pre-trained language models on the task when input exceeds the capacity of the models.


\begin{figure*}
\centering
\includegraphics[width=0.9\textwidth]{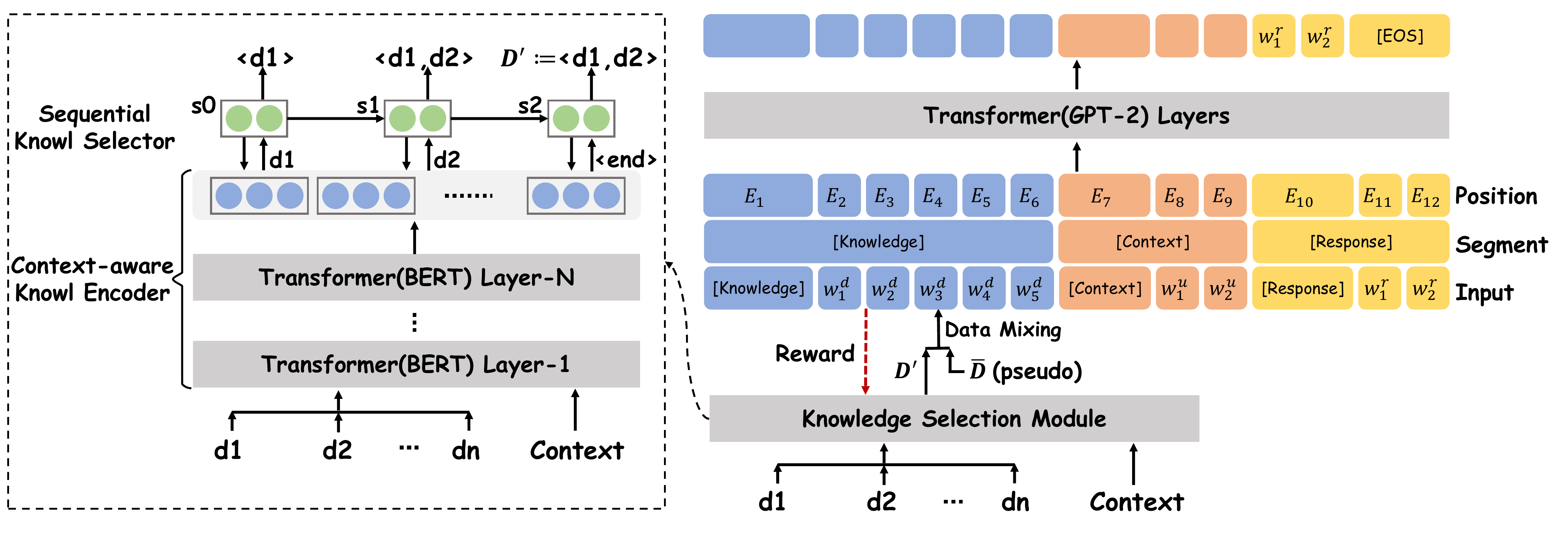}
\caption{Architecture of the proposed model. }
\label{fig:one}
\end{figure*}

\section{Preliminary}

\subsection{Problem Formalization}
Suppose that we have a dataset $\mathcal{D} = \{(U_i, D_i, r_i)\}_{i=1}^N$, where $\forall i \in \{1,\ldots,N\}$, $U_i$ is a dialogue context, $D_i$ is a document that contains relevant knowledge regarding to $U_i$, and $r_i$ is a response to $U_i$ based on $D_i$. The goal is to learn a generation model $P(r|U,D;\theta)$ ($\theta$ denotes the parameters of the model) from $\mathcal{D}$, and thus given a new dialogue context $U$ associated with a document $D$, one can generate a response $r$ following $P(r|U,D; \theta)$.

\subsection{Pre-trained Language Models}
We define $P(r|U,D; \theta)$ on the basis of GPT-2 from OpenAI \cite{radford2019language}. GPT-2 are transformer language models with a stack of masked multi-head self-attention layers, and are learned from large scale web text. 
To apply GPT-2 to the task of knowledge-grounded dialogue generation, we formulate the generation problem as
\begin{equation}
\begin{aligned}
P(r|U,D; \theta)&=P(r|g(U,D); \theta) \\
&=\prod_{t=1}^{l_r} P(r_{t}|g(U,D), r_{1:t-1}; \theta),
\end{aligned}
\end{equation}
where $g(U,D)$ tailors $U \cup D$ to meet the length constraint of a GPT-2 model as the input of generation, and $r_t$ refers to the $t$-th token of $r$ whose length is supposed to be $l_r$.  The problem then boils down to (1) how to define $g(U,D)$; and (2) how to fine-tune $\theta$ (and probably learn $g(U,D)$) with $\mathcal{D}$.

In this work, we assume that labels that indicate the ground-truth knowledge are not available, which is practical but makes the problem even more challenging. Since $D$ could be rather redundant with a lot of information irrelevant with the topic or the context of the conversation, simply truncating the concatenation of sentences of $U$ and $D$ as $g(U,D)$ may cut the relevant knowledge and introduce noise into response generation, which hurts the performance of the GPT-2 model, as will be demonstrated in the experiments. Therefore, we consider learning a $g(U,D)$ that can distill useful information from $D$ for the GPT-2 model, as will be elaborated in the next section.

\section{Approach}

Heading for learning a $g(U,D)$ for applying GPT-2 to the task of knowledge-grounded dialogue generation, we need to deal with several challenges: (1) how to model the correlation between a context and the external knowledge; (2) how to learn $g(U,D)$ when labels of ground-truth knowledge are absent; and (3) how to jointly optimize $g(U,D)$ and the GPT-2 model with $\mathcal{D}$, and thus the two can boost each other. Figure \ref{fig:one} illustrates the architecture of the model.  On the basis of the transformer architecture, the knowledge selection module is made up of a context-aware knowledge encoder and a sequential knowledge selector. The former captures interaction patterns between a context $U$ and each sentence in $D$ through a stack of self-attention layers, and the patterns are then fed to the latter to decode useful knowledge one sentence per step. Since human annotations are not accessible, the learning method begins with pseudo ground-truth constructed by making full use of responses, and optimization of $g(U,D)$ and optimization of the GPT-2 generation model are alternatively conducted with a reinforcement learning approach and a curriculum learning approach respectively.

\subsection{Context-Aware Knowledge Encoder}
We choose BERT \cite{devlin2018bert} as the backbone of the encoder. Thus, the encoder can take advantage of pre-training, and the multi-layer bi-directional attention mechanism in BERT allows a dialogue context and the associated knowledge to sufficiently interact with each other, resulting in context-aware knowledge representations. Specifically, let $U=(u_{1},\ldots, u_{n})$ and  $D=(d_{1}, \ldots, d_{m})$ be the context and the knowledge respectively, then we concatenate $\{u_{i}\}_{i=1}^{n}$ as $(w_{1}^u, \cdots, w_{l_u}^u)$ with $w_{i}^u$ the $i$-th word and $l_u$ the length of the sequence, and define the input of the encoder as $\mathcal{S}=(S_1,\ldots, S_m)$ with $S_i$ formulated as
\begin{equation} 
    S_i\!=[\mathrm{CLS}]\!w_{1}^u\!\ldots\!w_{l_u}^u\![\mathrm{SEP}]\!w_{i,1}^d\!\ldots\!w_{i,j}^d\!\ldots\!w_{i,l_d}^d\![\mathrm{SEP}],
\end{equation}
where $w_{i,j}^d$ refers to the $j$-th word of $d_i \in D$, and $l_d$ is the length of $d_i$.  Each $S_i \in \mathcal{S}$ passes through the stacked self-attention layers, and is finally represented as $e_{i} = \mathrm{CLS}(\mathrm{BERT}(S_i))$ where $\mathrm{BERT}(S_i)$ refers to the sequence of vectors from the last layer of the encoder and $\mathrm{CLS}(\cdot)$ is a function that returns the first vector of the sequence (i.e., the vector corresponding to the $[\mathrm{CLS}]$ token). The output of the encoder is given by $E=(e_{1}, \ldots, e_{m})$.

\subsection{Sequential Knowledge Selector}\label{KS}
With $E$ as input, the sequential knowledge selector determines a subset of $D$ (denoted as $D^{\prime}$) as the relevant knowledge and exploits $D^{\prime}$ to construct $g(U, D)$. Since there may exist one-to-many relations between a context and the relevant knowledge \cite{kim2020sequential}, the size of $D^{\prime}$ could vary from context to context. Therefore, we regard the construction of $D^{\prime}$ as a sequence prediction process in which $D^{\prime}$ starts from an empty set and gradually expands by adding one sentence from $D$ per step. By this means, the size of $D^{\prime}$ can also be viewed as a parameter and is dynamically determined according to the given context. Formally, we maintain a sequence of hidden states $\{s_t\}_{t=0}^{T_{U, D}}$ with the initial state $s_0$ a trainable parameter, and weight $\{d_i\}_{i=1}^m$ by an attention mechanism which can be formulated as  
\begin{equation}
\begin{aligned}
    &P(d_i | U, d_{j_{1:t-1}})=\exp (\alpha_{t, i}) / \sum_{i} \exp (\alpha_{t, i}) \\
    &\alpha_{t, i}=v^{\top}\tanh(W_{e}e_{i}+W_{s}s_{t}+b),
\label{sks}
\end{aligned}
\end{equation}

where $W_{e}$, $W_{s}$, $b$ and $v$ are trainable parameters. Then $d_{j_t}$ will be added to $D^{\prime}$ if $j_t = \operatorname{argmax}_{i\in \{1,\ldots,m\}}P(d_i | U, d_{j_{1:t-1}})$. After that, $s_{t+1}$ is calculated by
\begin{equation}
    s_{t+1}=\operatorname{LSTM}(e_{j_t}, s_{t})
\end{equation}

To determine $T_{U,D}$, we introduce a special embedding $e_{spe}$ into $E$, and terminate the prediction process if $e_{spe}$ is selected or an upper bound $T_{max}$ is reached. Finally, $g(U, D)$ is defined as the concatenation of the sentences in $U \cup D^{\prime}$.

\begin{algorithm*}
\small
\begin{algorithmic}[1]
    \State {\bfseries Input:} Training data $\mathcal{D}$, pre-trained GPT-2, initial curriculum rate $p_0$, exponential decay constant $\lambda$, maximum step $M$.
    \State Construct $\mathcal{D}_{K}$ and $\mathcal{D}_{G}$.
    \State Optimize $g(U, D)$ and GPT-2 using MLE on $\mathcal{D}_K$ and $\mathcal{D}_G$ respectively.
    \For {$m \gets 1 \mbox{~to~} M$}
        \State Sample a mini-batch $\{(U_i, D_i, r_i)\}$ from $\mathcal{D}$.
        \State Update the parameters of $g(U, D)$ based on Eq.\ref{eq:ks}. \Comment {the Reinforcement Step.}
        \State Sample $\{z_{i}\}$ from a Bernoulli distribution parameterized by $p$, where $p=p_{0} e^{-\lambda m}$.
        \State Update the parameters of the GPT-2 model based on Eq.\ref{eq:gpt2}. \Comment {the Curriculum Step.}
    \EndFor
    \State {\bfseries return}  $g(U, D)$ and GPT-2.
\end{algorithmic}
\caption{Optimization Algorithm}
\label{algo}
\end{algorithm*}

\subsection{Learning Method}
Learning a $g(U, D)$ without human annotations is not trivial. For example, in a recent paper \cite{kim2020sequential}, when human labels are removed, the accuracy of knowledge selection drops from $27$\% to $0.3$\%. Moreover, since knowledge selection and response generation are entangled, ideally we hope $g(U,D)$ and the GPT-2 model can enhance each other in learning. However, as the parameters of $g(U,D)$ are far from optimal at the early stage, it is very possible that noise from $g(U,D)$ will be fed to the GPT-2 model and then flows back to the learning procedure of $g(U,D)$, resulting in inferior models on both sides.
To cope with the challenges, we propose a joint optimization strategy with weak supervision as follows.
The learning algorithm is summarized in Algorithm \ref{algo}.

\paragraph{Pseudo Ground-Truth Construction.}

\label{sec:pseudo}
To alleviate error accumulation in joint optimization, we consider constructing weak supervision and utilize the signals to warm up the learning of $g(U, D)$ and the fine-tuning of GPT-2  beforehand. The intuition is that responses from humans carry clues to relevance of the knowledge candidates, and thus can be used to construct pseudo ground-truth. To be specific, we first sort $D = \{d_{t}\}_{t=1}^m$ in a descending order as $\{d_{j_t}\}_{t=1}^m$ according to $\{\operatorname{Sim}(d_t, r)\}_{t=1}^m$ where $\operatorname{Sim}(\cdot, \cdot)$ denotes a similarity function, and then build a subset of $D$ by
\begin{equation}
\begin{aligned}
    \bar{D} &= \{d_{j_1},\ldots, d_{j_{\bar{m}}}\}, \\
    \bar{m} &= \operatorname{argmax}_{t}(\operatorname{Sim}(d_{j_{1:t}}, r)),
\end{aligned}
\end{equation}
where $d_{j_{1:t}}$ refers to the concatenation of $\{d_{j_i}\}_{i=1}^t$. With $\bar{D}$, $g(U, D)$ and the GPT-2 model are optimized via maximum likelihood estimation (MLE) on $\mathcal{D}_{K} = \{(U_i, D_i, \bar{D}_i)\}_{i=1}^N$ and $\mathcal{D}_{G} = \{(U_i, \bar{D}_i, r_i)\}_{i=1}^N$ respectively.

\paragraph{Joint Optimization: the Reinforcement Step.}
We exploit the policy-gradient method \cite{sutton2000policy} to continue-train $g(U,D)$ by which $g(U,D)$ is further ``supervised'' by the GPT-2 model and is directly optimized for a target  metric (e.g., F1 in the experiments).  Specifically, we sample a $\tilde{D}$ according to $P(d_i | U, d_{j_{1:t-1}})$ (in Eq.\ref{sks}.)  under a termination criterion similar to $\bar{D}$ at each time step, and define the loss function as
\begin{equation} 
\begin{aligned}
    \mathcal{L}_K &= -\frac{1}{N}\sum_{i=1}^{N}\left(\tilde{R}_i\sum_{t=1}^{\left|\tilde{D}_i\right|}\log P(d_{i,j_t} | U_i, d_{i,j_{1:t-1}})\right), \\
    \tilde{R}_i &= R(\tilde{D}_i)- b,
\label{eq:ks}
\end{aligned}
\end{equation}
where $R(\tilde{D}_i)=\operatorname{Sim}(r'_i, r_i)$ with $r'_i$ the response generated by the GPT-2 model given $U_i$ and $\tilde{D}_i$, and $b=\sum_{i=1}^{N} R(\tilde{D}_{i}) / N$ is the baseline that is used to reduce the variance of gradient estimation\cite{clark2016deep}. We can see that minimizing $\mathcal{L}_K$ is equivalent to maximizing the conditional likelihood of $\tilde{D}_i$ if it obtains a higher reward than the baseline.

\paragraph{Joint Optimization: the Curriculum Step.}
Though $g(U,D)$ has been pre-trained with the pseudo ground-truth $\bar{D}$, the relevant knowledge provided by the model (i.e., $D^{\prime}$) may still be worse than $\bar{D}$ at the beginning of fine-tuning. Therefore, we mix $D^{\prime}$ and $\bar{D}$ and exploit a curriculum learning strategy to fine-tune the GPT-2 model where $D^{\prime}$ and $\bar{D}$ are regarded as hard materials and easy materials respectively and fine-tuning gradually moves from $\bar{D}$ to  $D^{\prime}$. Formally, the loss function for fine-tuning the GPT-2 model is defined by
\begin{equation} 
\begin{aligned}
    \mathcal{L}_G = &-\frac{1}{N} \sum_{i=1}^{N} \left(z_i \sum_{t=1}^{l_r} \log P(r_{i,t} | U_i, \bar{D}_i, r_{i,1:t-1}) \right.\\
    &\left.  +(1-z_i) \sum_{t=1}^{l_r} \log P(r_{i,t} | U_i, D_i^{\prime}, r_{i,1:t-1}) \right),
\label{eq:gpt2}
\end{aligned}
\end{equation}
where $\{z_i\}$ are sampled from a Bernoulli distribution parameterized by $p$. By gradually shrinking $p$, the generation model will be exposed to more hard materials with the learning procedure going on.

\section{Experiments}

We conduct experiments on Wizard of Wikipedia (Wizard) and CMU Document Grounded Conversations (CMU$\_$DoG) \cite{zhou2018dataset}.

\subsection{Datasets and Evaluation Metrics}
Both datasets are built with crowd-sourcing on Amazon Mechanical Turk, employ Wikipedia as the knowledge base, and are split into training sets, validation sets, and test sets by the data owners. Topics in Wizard cover a wide range ($1,365$ in total), and each conversation happens between a wizard who has access to the knowledge about a specific topic and an apprentice who is just eager to learn from the wizard about the topic. The test set is split into two subsets: Test Seen and Test Unseen. Test Seen contains new dialogues with topics appearing in the training set, while topics in Test Unseen never appear in the training set and the validation set. We follow \cite{dinan2018wizard} and conduct the pre-processing with the code published on ParlAI\footnote{\scriptsize\url{https://github.com/facebookresearch/ParlAI/blob/master/projects/wizard\_of\_wikipedia}}. Different from Wizard, CMU$\_$DoG focuses on movie domain, and besides wizard-apprentice conversations, the data also contain conversations between two workers who know the document and try to discuss the content in depth. To better compare with the baselines, we adopt the version shared at \url{https://github.com/lizekang/ITDD}. In both data, only the turns where knowledge is accessible are considered in response generation. More details are described in supplementary material.

We choose perplexity (PPL) of the ground-truth responses, BOW Embedding~\citep{liu2016not}, and unigram F1 \cite{dinan2018wizard} as metrics, where Embedding-based metrics are computed with an NLG evaluation open source available at  \url{https://github.com/Maluuba/nlg-eval}, and F1 is calculated with the code published at \url{https://github.com/facebookresearch/ParlAI/blob/master/parlai/core/metrics.py}. 

Besides automatic evaluation, we randomly sample $300$ examples from Test Seen, Test Unseen, and the test set of CMU$\_$DoG respectively, and recruit $3$ well-educated native speakers as annotators for human evaluation. To each annotator, an example is presented with a context, the associated external knowledge\footnote{For ease of labeling, only the ground-truth knowledge is shown to the annotators in Wizard.}, and model responses (top 1 in greedy search) that are randomly shuffled to hide their sources. The annotators then judge the quality of the responses from three aspects, including  \textit{fluency}, \textit{context coherence} and \textit{knowledge relevance}, and assign a score in  $\{0, 1, 2\}$ (representing ``bad'', ``fair'', and ``good'') to each response for each aspect. Each response receives $3$ scores per aspect, and the agreement among the annotators is measured via Fleiss' kappa \cite{fleiss1971measuring}.

\begin{table*}[h!]
\centering
\resizebox{1.0\linewidth}{!}{
\begin{tabular}{l|c|c|c|c|c|c|c|c|c|c}
\thickhline
\multicolumn{1}{c|}{\multirow{2}{*}{Models}} & \multicolumn{5}{c|}{Test Seen}           & \multicolumn{5}{c}{Test Unseen}        \\ \cline{2-11} 
\multicolumn{1}{c|}{}                        & PPL  & F1   & Average & Extrema & Greedy & PPL   & F1 & Average & Extrema & Greedy \\ \hline
TMN~\citep{dinan2018wizard}                  & 66.5 & 15.9 & 0.844   & 0.427   & 0.658  & 103.6 & 14.3 & 0.839 & 0.408   & 0.645      \\
ITDD~\citep{li2019incremental}               & 17.8 & 16.2 & 0.841   & 0.425   & 0.654  & 44.8  & 11.4 & 0.826 & 0.364 & 0.624      \\
SKT*~\citep{kim2020sequential}               & 52.0 & 19.3 & 0.846   & 0.440   & 0.665  & 81.4  & 16.1 & 0.839 & 0.418 & 0.652      \\
DRD~\citep{zhao2020low}                      & 19.4 & 19.3 & 0.852   & 0.452   & 0.674  & 23.0  & 17.9 & 0.849 & 0.439 & 0.664      \\ \hline
SKT+GPT-2*                                   & 17.6 & 20.3 & 0.866   & 0.460   & 0.679  & 23.7  & 17.8 & 0.860 & 0.437 & 0.664      \\
GPT-2$_{trunc}$                              & 14.6(2.2) & 18.7(0.7) & 0.864(0.002)   & 0.451(0.006)   & 0.674(0.004)  & 16.9(3.1)  & 18.3(0.6) & 0.862(0.002) & 0.444(0.005) & 0.668(0.003)      \\ \hline
KnowledGPT                                   & 19.2 & \textbf{22.0} & \textbf{0.872} & \textbf{0.463} & \textbf{0.682}  & 22.3  & \textbf{20.5} & \textbf{0.870} & 0.452 & \textbf{0.674}      \\ 
\thickhline
\end{tabular}
}
\caption{Evaluation results on Wizard. Models that leverage human labels are marked with *. Numbers in bold mean that the improvement to the best baseline is statistically significant (t-test with $p$-value $<$ 0.01).}
\label{tab:wizard_exp}
\end{table*}

\begin{table}[h!]
\centering
\resizebox{1.0\linewidth}{!}{
\begin{tabular}{l|c|c|c|c|c}
\thickhline
\multicolumn{1}{c|}{Models}       & PPL  & F1   & Average & Extrema & Greedy \\ \hline
TMN~\citep{dinan2018wizard}       & 75.2 & 9.9  & 0.789   & 0.399   & 0.615  \\
ITDD~\citep{li2019incremental}    & 26.0 & 10.4 & 0.748   & 0.390   & 0.587  \\
DRD~\citep{zhao2020low}           & 46.1 & 10.8 & 0.791   & 0.406   & 0.613  \\
\hline
GPT-2$_{trunc}$                   & 18.6 & 10.8  & 0.730  & 0.419   & 0.597  \\ \hline
KnowledGPT                        & 20.6 & \textbf{13.5}  & \textbf{0.837} & \textbf{0.437} & \textbf{0.654}     \\ 
\thickhline
\end{tabular}
}
\caption{Evaluation results on CMU$\_$DoG. Numbers in bold mean that the improvement to the best baseline is statistically significant (t-test with $p$-value $<$ 0.01).}
\label{tab:cmudog_exp}
\end{table}

\begin{table*}[]
\resizebox{1.0\linewidth}{!}{
\begin{tabular}{l|c|c|c|c|c|c|c|c|c|c|c|c}
\thickhline
\multicolumn{1}{c|}{\multirow{3}{*}{Models}} & \multicolumn{8}{c|}{Wizard} & \multicolumn{4}{c}{\multirow{2}{*}{CMU$\_$DoG}} \\ \cline{2-9}
\multicolumn{1}{c|}{} & \multicolumn{4}{c|}{Test Seen} & \multicolumn{4}{c|}{Test Unseen} & \multicolumn{4}{c}{} \\ \cline{2-13} 
\multicolumn{1}{c|}{}                        & Fluency & \begin{tabular}[c]{@{}c@{}}Context\\ Coherence\end{tabular} & \begin{tabular}[c]{@{}c@{}}Knowledge\\ Relevance\end{tabular} & Kappa & Fluency & \begin{tabular}[c]{@{}c@{}}Context\\ Coherence\end{tabular} & \begin{tabular}[c]{@{}c@{}}Knowledge\\ Relevance\end{tabular} & Kappa & Fluency & \begin{tabular}[c]{@{}c@{}}Context\\ Coherence\end{tabular} & \begin{tabular}[c]{@{}c@{}}Knowledge\\ Relevance\end{tabular} & Kappa \\ \hline
DRD             & 1.71 & 1.50 & 1.26 & 0.67 & 1.64 & 1.44 & 1.18 & 0.69 & 1.58 & 1.48 & 1.07 & 0.60  \\ \hline
GPT-2$_{trunc}$ & 1.86 & 1.54 & 1.22 & 0.71 & 1.84 & 1.47 & 1.20 & 0.59 & 1.83 & 1.58 & 1.06 & 0.64  \\ \hline
KnowledGPT      & 1.89 & 1.67 & 1.71 & 0.70 & 1.88 & 1.60 & 1.68 & 0.73 & 1.83 & 1.65 & 1.50 & 0.77  \\ 
\thickhline
\end{tabular}
}
\caption{Human evaluation results on Wizard and CMU$\_$DoG.}
\label{tab:human}
\end{table*}

\subsection{Baselines}
The following models are selected as baselines:

\textbf{Transformer Memory Network (TMN):} the model proposed in \cite{dinan2018wizard} along with the release of the Wizard data. We implement it using the code shared at {\url{https://github.com/facebookresearch/ParlAI/blob/master/projects/wizard_of_wikipedia}}.

\textbf{Incremental Transformer with Deliberation Decoder (ITDD):} a transformer-based model \cite{li2019incremental} that incrementally encodes multi-turn dialogues and knowledge and decodes responses with a deliberation technique. We implement it using the code shared at \url{https://github.com/lizekang/ITDD}.

\textbf{Sequential Knowledge Transformer (SKT):} a sequential latent variable model with state-of-the-art performance on knowledge selection published in a very recent paper \citep{kim2020sequential}. Since human labels that indicate ground-truth knowledge are crucial to the performance of the model, we only involve it as a baseline on the Wizard data. The model is implemented with the code shared at   \url{https://github.com/bckim92/sequential-knowledge-transformer}.

\textbf{Disentangled Response Decoder (DRD):} a model that tackles the low-resource challenge with pre-training techniques \cite{zhao2020low}. We choose the one in which all parameters are fine-tuned with the full training data after pre-training as the baseline, since such a configuration results in state-of-the-art performance on Wizard, as reported in \cite{zhao2020low}.

We name our model \textbf{KnowledGPT}. Besides the baselines described above, the following pre-trained models are also included in comparison in order to have a thorough understanding towards the proposed method: (1) \textbf{GPT-2$_{trunc}$}. We concatenate a context and the associated knowledge as a long document, and then truncate the document to meet the length constraint of the GPT-2 model. This is to check if the simple heuristics work for the task. Note that in Wizard, we randomly mix the ground-truth knowledge with others and repeat the procedure $8$ times. The means with standard deviation (i.e., numbers in ``( )'') are reported to remove randomness; 
and (2) \textbf{SKT+GPT-2}. We feed the candidate selected by SKT to GPT-2 for response generation. This is to examine if we can simply replace the proposed knowledge selection module as well as the learning approach with an off-the-shelf knowledge selection model. Similar to SKT, the comparison is only conducted on Wizard.

\subsection{Implementation Details}

In both Wizard and CMU\_DoG, we set the hidden size and the number of layers of the sequential knowledge selector as $256$ and $1$ respectively. $T_{max}$ for $D^{\prime}$ is set as $1$ in Wizard, and $2$ in CMU\_DoG. We choose BERT (110M) and GPT-2 (117M) as the pre-trained language models in KnowledGPT, and implement the models with the code in \url{https://github.com/huggingface/transformers}. We employ greedy search in response decoding. All models are learned with Adam \cite{kingma2014adam} optimizer with $\beta_1=0.9$ and $\beta_2=0.999$. In warming up, we define $\operatorname{Sim}(\cdot,\cdot)$ as unigram F1, and optimize $g(U,D)$ and the GPT-2 model with the pseudo ground-truth for $1000$ steps with a batch size of $64$. In joint optimization, the batch size is set as $128$, and the learning rates for $g(U,D)$ and GPT-2 are set as $5e-6$ and $5e-5$ respectively. The learning rate will be halved if there is no improvement in terms of PPL on the validation sets. The parameter $p$ of the Bernoulli distribution in the curriculum step is initially set as $1.0$ and anneals with a rate of $1e-5$. Early stopping on validation is adopted as a regularization strategy.

\begin{table*}[]
\resizebox{1.0\linewidth}{!}{
\begin{tabular}{l|c|c|c|c|c|c|c|c|c|c|c|c|c|c|c}
\thickhline
\multicolumn{1}{c|}{\multirow{3}{*}{Models}} & \multicolumn{10}{c|}{Wizard} & \multicolumn{5}{c}{\multirow{2}{*}{CMU$\_$DoG}} \\ \cline{2-11} \multicolumn{1}{c|}{} & \multicolumn{5}{c|}{Test Seen} & \multicolumn{5}{c|}{Test Unseen} & \multicolumn{5}{c}{} \\ \cline{2-16} 
\multicolumn{1}{c|}{} & PPL  & F1   & Average & Extrema & Greedy & PPL  & F1   & Average & Extrema & Greedy & PPL    & F1     & Average   & Extrema  & Greedy  \\ \hline
KnowledGPT  & 19.2 & 22.0 & 0.872   & 0.463   & 0.682  & 22.3 & 20.5 & 0.870   & 0.452   & 0.674  & 20.6   & 13.5   & 0.837     & 0.437    & 0.654   \\ \hline
-pseudo     & 22.3 & 18.3 & 0.857   & 0.436   & 0.662  & 24.1 & 17.9 & 0.854   & 0.430   & 0.655  & 23.2   & 12.9   & 0.815     & 0.440    & 0.639   \\
-joint & 20.0 & 20.4 & 0.863   & 0.457   & 0.675  & 21.8 & 19.5 & 0.861  & 0.451   & 0.669  & 22.6   & 11.7   & 0.806     & 0.438    & 0.635  \\
-curriculum  & 19.4 & 21.2 & 0.867   & 0.457   & 0.677  & 21.5 & 20.3 & 0.866   & 0.451   & 0.672  & 21.9   & 12.4   & 0.816     & 0.443    & 0.644   \\
-reinforcement  & 19.4 & 21.3 & 0.866   & 0.459   & 0.677  & 21.9 & 20.2 & 0.863  & 0.449   & 0.670  & 20.3   & 12.6   & 0.817     & 0.437    & 0.643  \\
\thickhline
\end{tabular}
}
\caption{Ablation study on Wizard and CMU$\_$DoG}
\label{tab:abl}
\end{table*}

\subsection{Evaluation Results}

Table \ref{tab:wizard_exp} and Table \ref{tab:cmudog_exp} report evaluation results on Wizard and CMU$\_$DoG respectively. KnowledGPT achieves new state-of-the-art on most metrics in both datasets, which demonstrates the effectiveness of large-scale pre-trained language models on the task of knowledge-grounded dialogue generation. 
GPT-2$_{trunc}$ is worse than KnowledGPT, due to (1) knowledge loss: we find that in $53\%$ test examples (Test Seen+Test Unseen), the ground-truth knowledge is cut. In this case, GPT-2$_{trunc}$ only relies on the context, the related knowledge in other candidates (thanks to the one-to-many relations between a context and knowledge), and the knowledge packed in the parameters of GPT-2 for responding, which explains the comparable performance with SKT and DRD; and (2) noisy input: even though the ground-truth knowledge is kept, the redundant and irrelevant information in the knowledge candidates are still harmful. Evidence is that GPT-2$_{trunc}$ is worse than KnowledGPT on CMU\_DoG even though we do not cut anything on the knowledge (the maximum length of the knowledge input is $502$, and thus is within the constraint of GPT-2). KnowledGPT also outperforms SKT+GPT-2 on Wizard, because (1) KnowledGPT is more accurate than SKT on knowledge selection, even though it does not leverage any human annotations in learning. In fact, the accuracy scores of knowledge selection for SKT are $26.8$ and $18.3$ on Test Seen and Test Unseen respectively, while the two numbers are $28.0$ and $25.4$ respectively for KnowledGPT; and 
(2) in KnowledGPT, knowledge selection and response generation are jointly optimized.

Table \ref{tab:human} shows human evaluation results. While the three models are comparable on \textit{fluency}, KnowledGPT is superior to the others on both \textit{context coherence} and \textit{knowledge relevance}, which is consistent with the results on automatic metrics. All kappa values are no less than $0.6$, indicating substantial agreement among the annotators. We present a case study in supplementary material.

\subsection{Discussions}

\textbf{Ablation study.} To understand the impact of the learning strategies on model performance, we compare the full KnowledGPT with the following variants: (1) \textit{-pseudo}: the warming up stage is removed; (2) \textit{-joint}: the joint optimization stage is removed; (3) \textit{-reinforcement}: $g(U,D)$ is fixed after it is optimized with MLE on $\mathcal{D}_{K}$; and (4) \textit{-curriculum}:  GPT-2 is fixed after it is optimized with MLE on $\mathcal{D}_{G}$.
Table \ref{tab:abl} reports the evaluation results. We can conclude that 
(1) the pseudo ground-truth plays a crucial role in Wizard, as removing the step causes dramatic performance drop. This is because in Wizard, there is a strong correlation between the knowledge and human responses. The results indicate that though the pseudo ground-truth is constructed with heuristics, it still contains valuable information and thus allows the following joint optimization to start from a good point. On the other hand, in CMU\_DoG, the crowd-workers do not refer to the external knowledge as much as those workers do in Wizard when they form the responses; 
(2) the reinforcement step and curriculum step are useful because the reinforcement step allows the knowledge selection module to make better use of GPT-2's feedback, and through the curriculum step GPT-2 can take advantage of the output of knowledge selection module progressively;
(3) joint optimization is meaningful, as removing this stage results in performance drop.

\begin{table}[]
\centering
\resizebox{0.85\linewidth}{!}{
\begin{tabular}{l|c|c|c|c|c|c}
\thickhline
\multicolumn{1}{c|}{\multirow{3}{*}{Models}} & \multicolumn{4}{c|}{Wizard} & \multicolumn{2}{c}{\multirow{2}{*}{CMU\_DoG}} \\ \cline{2-5}
\multicolumn{1}{c|}{} & \multicolumn{2}{c|}{Test Seen} & \multicolumn{2}{c|}{Test Unseen} & \multicolumn{2}{c}{} \\ \cline{2-7} 
\multicolumn{1}{c|}{} & PPL & F1 & PPL & F1 & PPL & F1 \\ \hline
T$_{max}$=1 & 19.2 & 22.0 & 22.3 & 20.5 & 20.6 & 12.6 \\ \hline
T$_{max}$=2 & 18.2 & 21.3 & 21.0 & 20.3 & 20.6 & 13.5 \\ \hline
T$_{max}$=3 & 17.2 & 21.1 & 20.2 & 20.3 & 19.7 & 11.2 \\ 
\thickhline
\end{tabular}
}
\caption{Performance of KnowledGPT under different $T_{max}$s.}
\label{tab:knowl_len}
\end{table}

\textbf{Impact of  $T_{max}$ (i.e., the upper bound in knowledge selection).} Besides the learning strategies, we are also curious about how $T_{max}$, as part of the termination criterion in knowledge selection described at the end of Section \ref{KS}, influences the performance of KnowledGPT. To this end, we vary the value of $T_{max}$ in $\{1,2,3\}$ and report the evaluation results in Table \ref{tab:knowl_len}. The larger $T_{max}$ is, the more chances KnowledGPT has to involve the ground-truth candidate into generation, and the lower PPL is. This also explains why the PPL of GPT-2$_{trunc}$ is lower than that of KnowledGPT in Table \ref{tab:wizard_exp} and Table \ref{tab:cmudog_exp}. On the other hand, a larger $T_{max}$ also means more noise in generation. That is why when $T_{max}$ exceeds a value, F1 begins to drop.

\section{Conclusions}
We apply large-scaled pre-trained language models to the task of knowledge-grounded dialogue generation. To this end, we devise a knowledge selection module, and propose an unsupervised approach to jointly optimizing knowledge selection and response generation. Evaluation results on two benchmarks indicate that our model can significantly outperform state-of-the-art methods.

\subsection*{Acknowledgments}
We would like to thank the reviewers for their constructive comments. This work was supported by the National Key Research and Development Program of China (No. 2020AAA0105200), the National Science Foundation of China (NSFC No. 61876196 and NSFC No. 61672058). Rui Yan was sponsored as the young fellow of Beijing Academy of Artificial Intelligence (BAAI). Rui Yan is the corresponding author.


\bibliographystyle{acl_natbib}
\bibliography{anthology,emnlp2020}

\clearpage
\appendix

\section{Details of Datasets}\label{app:dataset}

Table \ref{tbl:stat} reports the statistics of the Wizard data and the CMU$\_$DoG data. 

\begin{table}[H]
\centering
\resizebox{1.0\linewidth}{!}{
\begin{tabular}{l|c|c|c|c|c|c|c}
\thickhline
\multirow{2}{*}{}          & \multicolumn{4}{c|}{Wizard of Wikipedia}   & \multicolumn{3}{c}{CMU$\_$DoG} \\ \cline{2-8}
                           & Train   & Valid  & Test Seen & Test Unseen & Train    & Valid  & Test    \\ \hline
$\#$ Utterances       & 166,787 & 17,715 & 8,715     & 8,782       & 74,717   & 4,993  & 13,646  \\ \hline
$\#$ Conversations    & 18,430  & 1,948  & 965       & 968         & 3,373    & 229    & 619     \\ \hline
$\#$ Topics/Documents & 1,247   & 599    & 533       & 58          & 30       & 30     & 30      \\ \hline
Avg. $\#$ of Turns & 9.0     & 9.1    & 9.0       & 9.1         & 22.2     & 21.8   & 22.0    \\
\thickhline
\end{tabular}
}
\caption{Statistics of the two datasets.}
\label{tbl:stat}
\end{table}

\section{Comparison with DialoGPT}
We compare KnowledGPT and with DialoGPT  in order to learn if a pre-trained generation model with state-of-the-art performance on open domain dialogues is already good enough when it is fine-tuned with knowledge-grounded dialogues. We discard the associated knowledge and fine-tune DialoGPT on the knowledge-grounded dialogues. We choose the model trained from OpenAI GPT-2 with $345$M parameters, as it shows the best performance in the evaluation in the original paper. The model is implemented based on the code shared at \url{https://github.com/microsoft/DialoGPT}. 

Table \ref{tab:dialogpt} shows the results, indicating that external knowledge is necessary even though one has exploited a powerful pre-trained language model for dialogue generation. In CMU$\_$DoG the gap between DialoGPT and KnowledGPT is narrowed because about 35\% of the conversation has a weak correlation with the document (e.g. BLEU $<$ 0.1).

\begin{table}[H]
\resizebox{1.0\linewidth}{!}{
\begin{tabular}{c|c|c|c|c|c|c}
\thickhline
\multirow{3}{*}{Models} & \multicolumn{4}{c|}{Wizard} & \multicolumn{2}{c}{\multirow{2}{*}{CMU$\_$DoG}} \\ \cline{2-5}
 & \multicolumn{2}{c|}{Test Seen} & \multicolumn{2}{c|}{Test Unseen} & \multicolumn{2}{c}{} \\ \cline{2-7} 
 & PPL  & F1   & PPL  & F1   & PPL  & F1  \\ \hline
\multicolumn{1}{l|}{DialoGPT}   & 16.0 & 17.9 & 20.0 & 16.8 & 16.9 & 12.3 \\ \hline
\multicolumn{1}{l|}{KnowledGPT} & 19.2 & 22.0 & 22.3 & 20.5 & 20.6 & 13.5 \\ \thickhline
\end{tabular}
}
\caption{Comparison with DialoGPT on Wizard and CMU$\_$DoG}
\label{tab:dialogpt}
\end{table}

\section{Impact of Maximum Tokens of GPT-2}

To further justify our claims on why GPT-2$_{trunc}$ is worse than KnowledGPT, we keep the ground-truth knowledge in the input sequence of GPT-2 and gradually increase the constraint of the maximum number of tokens on Wizard. As the maximum token limit increases, more irrelevant knowledge is introduced. Note that in practice, one has no way to perfectly locate the ground-truth, and this experiment is only to provide more insights to GPT-2$_{trunc}$. Table \ref{tab:involved} shows the performance of GPT-2$_{trunc}$ with the increase of the maximum number of tokens where Ground-truth Percentage indicates the percentage of ground-truth in the input knowledge. First, when the ground-truth is forced to be kept, GPT-2$_{trunc}$ is always better than the one where the ground-truth is randomly mixed with other candidates and bears the risk to be cut. This echoes our claim that knowledge loss is one of the reasons for the poor performance of  GPT-2$_{trunc}$ used with the practical setting. Second, even if ground-truth is retained, once more noise is introduced, the performance of GPT-2$_{trunc}$ will become worse. When the length is limited to $128$ tokens, the PPL of the model is not good, mainly because under this limitation, the input sequence of some cases only contains the dialogue context and response.

\begin{table}[]
\resizebox{1.0\linewidth}{!}{
\begin{tabular}{l|c|c|c|c|c}
\thickhline
\multicolumn{1}{c|}{\multirow{2}{*}{Maximum Tokens}} & \multicolumn{2}{c|}{Test Seen} & \multicolumn{2}{c|}{Test Unseen} & \multirow{2}{*}{\begin{tabular}[c]{@{}c@{}}Ground-truth\\ Percentage\end{tabular}} \\ \cline{2-5}
\multicolumn{1}{c|}{} & PPL            & F1            & PPL             & F1             & \\ \hline
128  & 10.8 & 30.9 & 11.6 & 30.4 & 62.3\% \\ \hline
256  & 9.3  & 25.6 & 10.0 & 24.6 & 20.3\% \\ \hline
512  & 9.7  & 21.8 & 10.5 & 21.2 & 8.5\%  \\ \hline
768  & 10.1 & 20.6 & 10.7 & 20.2 & 5.5\% \\ \hline
1024 & 10.7 & 19.7 & 11.3 & 19.4 & 4.1\% \\ 
\thickhline
\end{tabular}
}
\caption{Performance of GPT-2$_{trunc}$ under different maximum tokens with ground-truth knowledge involved.}
\label{tab:involved}
\end{table}

\section{Impact of the Size of GPT-2} 

We further check if the performance of KnowledGPT can be further improved when the GPT-2 model is replaced with a larger one. Table \ref{tab:gpt_size} shows the results. Though GPT-2 (345M) can further reduce PPL, it does not bring significant improvement to F1 over GPT-2 (117M), probably because the larger model can not provide more accurate feedback to the knowledge selection module in learning.  Therefore, to balance efficacy and cost, GPT-2 (117M) is still favored in practice.  

\begin{table}[]
\centering
\resizebox{1.0\linewidth}{!}{
\begin{tabular}{c|c|c|c|c|c|c}
\thickhline
\multirow{3}{*}{Models} & \multicolumn{4}{c|}{Wizard of Wikipedia} & \multicolumn{2}{c}{\multirow{2}{*}{CMUDoG}} \\ \cline{2-5}
& \multicolumn{2}{c|}{Test Seen} & \multicolumn{2}{c|}{Test Unseen} & \multicolumn{2}{c}{} \\ \cline{2-7} 
                                      & PPL  & F1   & PPL  & F1   & PPL  & F1   \\ \hline
\multicolumn{1}{l|}{KnowledGPT (117M)}       & 19.2 & 22.0 & 22.3 & 20.5 & 20.6 & 13.5 \\ \hline
\multicolumn{1}{l|}{KnowledGPT (345M)} & 16.1 & 22.0 & 17.9 & 20.6 & 18.1 & 13.4 \\ 
\thickhline
\end{tabular}
}
\caption{Performance of  KnowledGPT under different sizes of GPT-2.}
\label{tab:gpt_size}
\end{table}

\section{Case Study}
Table \ref{tab:case1} and Table \ref{tab:case2} show the examples from Test Seen and Test Unseen of Wizard, each example contains the dialogue context and the background knowledge which is retrieved from Wikipedia given the last two turns of dialogue and the original topic. We can see that KnowledGPT can locate the knowledge more accurately due to its knowledge selection module and reinforcement learning, and make better use of the associated knowledge with the help of curriculum learning.

\begin{table*}[]
\resizebox{1.0\linewidth}{!}{
\begin{tabular}{cl}
\hline
\multicolumn{2}{c}{Knowledge (Topic: Cinematography)} \\ \hline
\multicolumn{2}{l}{...}  \\
\multicolumn{2}{p{800pt}}{Cinematography (also called "direction of photography") is the science or art of motion-picture photography by recording light or other electromagnetic radiation, either electronically by means of an image sensor, or chemically by means of a light-sensitive material such as film stock.} \\
\multicolumn{2}{p{800pt}}{Typically, a lens is used to repeatedly focus the light reflected from objects into real images on the light-sensitive surface inside a camera during a questioned exposure, creating multiple images.} \\
\multicolumn{2}{p{800pt}}{With an electronic image sensor, this produces an electrical charge at each pixel, which is electronically processed and stored in a video file for subsequent display or processing.} \\
\multicolumn{2}{p{800pt}}{The result with photographic emulsion is a series of invisible latent images on the film stock, which are later chemically "developed" into a visible image.} \\
\multicolumn{2}{p{800pt}}{The images on the film stock are played back at a rapid speed and projected onto a screen, creating the illusion of motion.} \\
\multicolumn{2}{p{800pt}}{Cinematography finds uses in many fields of science and business as well as for entertainment purposes and mass communication.} \\
\multicolumn{2}{p{800pt}}{The word "cinematography" was created from the greek words ("kinema"), meaning "movement, motion" and ("graphein") meaning "to record", together meaning "recording motion."} \\
\multicolumn{2}{l}{...}  \\ \hline
\multicolumn{2}{c}{Context}  \\ \hline
\multicolumn{1}{c|}{B}  & Thinking about taking a class on cinematography, the direction of photography .  \\
\multicolumn{1}{c|}{A} & I have always wanted to learn that .  \\
\multicolumn{1}{c|}{B} & Me too, I have learned its the science and or art of recording motion picture photography . \\
\multicolumn{1}{c|}{A}  & That is cool. did you enjoy it . \\ \hline
\multicolumn{2}{c}{Response} \\ \hline
\multicolumn{1}{c|}{TMN} & I think it is a great idea . It is a science fiction film that combines computer and computer science .  \\ \hline
\multicolumn{1}{c|}{ITDD} & I did . It was developed in the late 19th century . \\ \hline
\multicolumn{1}{c|}{SKT} & Yes , I did . I loved it . It was used to focus the light reflected from objects into real images . \\ \hline
\multicolumn{1}{c|}{DRD} & I did enjoy it . It is also called `` direction of photography '' . \\ \hline
\multicolumn{1}{c|}{SKT+GPT-2}  & I did! I love how it's so easy to focus light reflected from objects into real images on the light-sensitive surface inside a camera. \\ \hline
\multicolumn{1}{c|}{GPT-2$_{trunc}$} & Yes, I did, I love the fact that it is the science and art of recording motion picture photography \\ \hline
\multicolumn{1}{c|}{KnowledGPT}  & Yes, I did, I love cinematography, its used in many fields of science and business as well as for entertainment purposes and mass communication \\ \hline
\multicolumn{1}{c|}{Human} & Yes and cinematography skills can also be used in mass marketing and other fields as well \\
\hline
\end{tabular}
}
\caption{A case from Test Seen of Wizard of Wikipedia}
\label{tab:case1}
\end{table*}

\begin{table*}[]
\resizebox{1.0\linewidth}{!}{
\begin{tabular}{cl}
\hline
\multicolumn{2}{c}{Knowledge(Topic: Green)} \\ \hline
\multicolumn{2}{l}{...} \\
\multicolumn{2}{p{800pt}}{Green is the color between blue and yellow on the visible spectrum.} \\
\multicolumn{2}{p{800pt}}{It is evoked by light which has a dominant wavelength of roughly 495570 nm.} \\
\multicolumn{2}{p{800pt}}{In subtractive color systems, used in painting and color printing, it is created by a combination of yellow and blue, or yellow and cyan; in the rgb color model, used on television and computer screens, it is one of the additive primary colors, along with red and blue, which are mixed in different combinations to create all other colors.} \\
\multicolumn{2}{p{800pt}}{By far the largest contributor to green in nature is chlorophyll, the chemical by which plants photosynthesize and convert sunlight into chemical energy.} \\
\multicolumn{2}{p{800pt}}{Many creatures have adapted to their green environments by taking on a green hue themselves as camouflage.} \\
\multicolumn{2}{p{800pt}}{Several minerals have a green color, including the emerald, which is colored green by its chromium content.} \\
\multicolumn{2}{p{800pt}}{In the middle ages and renaissance, green was the color commonly associated with merchants, bankers and the gentry, while red was reserved for the nobility.} \\
\multicolumn{2}{p{800pt}}{For this reason, the costume of the "mona lisa" by leonardo da vinci and the benches in the british house of commons are green while those in the house of lords are red.} \\
\multicolumn{2}{p{800pt}}{It also has a long historical tradition as the color of ireland and of gaelic culture.} \\
\multicolumn{2}{l}{...} \\ \hline
\multicolumn{2}{c}{Context} \\ \hline
\multicolumn{1}{c|}{B} & Green is my favorite color. Many minerals are a green color, including emeralds. I like emeralds too! \\
\multicolumn{1}{c|}{A} & I like green too. My eyes are green. \\
\multicolumn{1}{c|}{B} & So are mine! In gaelic cultures green is very popular, with a long historical tradition. It is known as the color of ireland. \\
\multicolumn{1}{c|}{A} & Yes I do know it is a very popular color in ireland. \\ \hline
\multicolumn{2}{c}{Response} \\ \hline
\multicolumn{1}{c|}{TMN} & I think it ' s a very interesting subject . It ' s a culture of culture. \\ \hline
\multicolumn{1}{c|}{ITDD} & Yes , it is one of the primary colors . \\ \hline
\multicolumn{1}{c|}{SKT} & It is a popular color in the us \\ \hline
\multicolumn{1}{c|}{DRD} & It is common in nature , especially in plants . \\ \hline
\multicolumn{1}{c|}{SKT+GPT-2} & Yes, it is very popular in gaelic culture. \\ \hline
\multicolumn{1}{c|}{GPT-2$_{trunc}$} & It is! It is also the second largest island in the british isles. \\ \hline
\multicolumn{1}{c|}{KnowledGPT} & Yes, it is! In the middle ages and renaissance, green was the color commonly associated with merchants, bankers and the gentry.\\ \hline
\multicolumn{1}{c|}{Human} & Back in the middle ages green was often associated with merchants and bankers. Red was nobility's color of choice. \\
\hline
\end{tabular}
}
\caption{A case from Test Uneen of Wizard of Wikipedia}
\label{tab:case2}
\end{table*}

\end{document}